# High Performance Novel Skin Segmentation Algorithm for Images With Complex Background

Mohammad Reza Mahmoodi

Department of Electrical and Computer Engineering, Isfahan University of Technology, 8415683111  Isfahan, Iran
Email addresses: mr.mahmoodi@ec.iut.ac.ir

*Abstract*—**Skin Segmentation is widely used in biometric applications such as face detection, face recognition, face tracking, and hand gesture recognition. However, several challenges such as nonlinear illumination, equipment effects, personal interferences, ethnicity variations, etc., are involved in detection process that result in the inefficiency of color based methods. Even though many ideas have already been proposed, the problem has not been satisfactorily solved yet. This paper introduces a technique that addresses some limitations of the previous works. The proposed algorithm consists of three main steps including initial seed generation of skin map, Otsu segmentation in color images, and finally a two-stage diffusion. The initial seed of skin pixels is provided based on the idea of ternary image as there are certain pixels in images which are associated to human complexion with very high probability. The Otsu segmentation is performed on several color channels in order to identify homogeneous regions. The result accompanying with the edge map of the image is utilized in two consecutive diffusion steps in order to annex initially unidentified skin pixels to the seed. Both quantitative and qualitative results demonstrate the effectiveness of the proposed system in compare with the state-of-the-art works.**

I. INTRODUCTION

Skin classification is the act of separating skin and non-skin pixels in an arbitrary image. Skin segmentation algorithm is a major concern in many number of applications including face detection [1,2], content base retrieval [3], identifying and indexing multimedia information [4], gesture recognition [5,6], robotics [7], driver drowsiness monitoring [8,9], sign language recognition [10,11], gaming interfaces [12] and human computer interaction [13,14], and filtering objectionable URLs.

Considering the last one for example, recently, problems on the lack of regulation over available Internet information have become esteemed [15]. Explosion of online information influences every facet of human life and freedom access makes the World-Wide Web as the most popular place where people obtain, deliver and exchange information. But, this unconstrained information also brings reverse effects such as accessing illegal digital contraband. Generally, three different solutions have been addressed for this problem. Not the best but the mostly used technique is based on contextual keyword pattern matching technology that categorizes URLs by means of checking contexts of web pages and then traps the websites assorted as the obscene [16,17]. Though this method can filter out a mass of obscene websites, it is not capable of dealing with images. As the second approach, collecting and blocking objectionable websites based on the web address is not efficient as great number of new pages is created each day [18,19]. Content based techniques [20,21] which analyze add-up of web pages are designed based on skin color detectors. Here, the performance of filters will be dependent on the performance of skin classifiers. But, skin detectors have still long way to be acceptive and robust in dealing with challenges in this and also in many other applications. Hence, there is an immediate exigency on designing an efficient, robust skin detector to be used in such applications. The aim of this paper is to leverage an efficient skin classifier for different applications.

Skin detection is the process of classifying image's pixels into two categories. The first one includes pixels which belong to a human skin such as a face, hand or any part of human body, and the other one contains pixels which are not associated with human skin and it is not sensitive to the changes of posture and facial expression. Skin is invariant against rotation [22,23], geometrics [24], and is stable against partial occlusion [22,23], scaling and shape and it is somewhat person independent. Though skin segmentation task has several appealing features mentioned above, there are some synthetic and non-synthetic challenges effective in degrading the performance of classifiers. They include uneven, inconsistent and nonlinear illumination, complex, pseudo skin background, imaging equipment and camera dependency [23], and individual and intra-personal characteristics [25]. These challenges are mostly related to the algorithms which are based on visible spectrum imaging. For infrared systems, expensive systems, tedious setup procedures are two most common problems [23].

The color cannot deal with so called challenges mainly due to its high level of sensitivity particularly to illumination changes. Combination of color based processing and spatial analysis will enhance the performance significantly. Accordingly, a novel method based on both color and spatial analysis is exploited in this paper. A color processing is performed in order to produce an initial seed image for further processing. Spatial analysis is also combined with another color processing scheme i.e. Otsu segmentation method to boost the performance of the classifier. And diffusion process, based on several factors, from each pixel to other pixels in the image is performed to have skin pixels which are not extracted at the beginning be annexed into the final results.

The remainder of the paper is organized as follows. In section 2, some previous related works in the literature is briefly explained. In section 3, the proposed system is presented. Experimental results are illustrated in section 4, and finally, conclusion is provided in section 5.

## II. PREVIOUS WORKS

Different skin modeling techniques have been discussed in literature, each of them leaning on specific feature or cluster modeling scheme. In this section, most existing techniques in segmentation of skin pixels are elucidated. Fig. 1 shows the classification of skin detection methods.

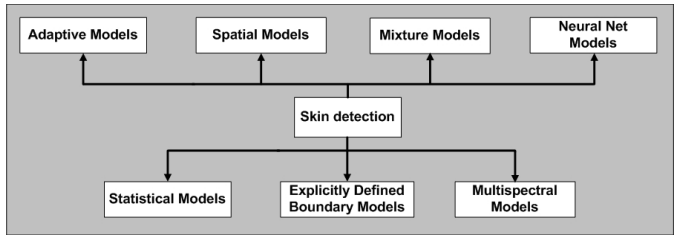

Fig. 1.  Skin detection methods

The principle difference in representation of human skin color is its intensity (brightness) rather than its chrominance (color) [26]. This is simply explainable by examining histogram of skin pixels in different color spaces. This observation has been a leading factor in developing a very common and simple approach in skin classification.

Explicitly defined methods are based on a set of rules derived from the skin locus in 2D or 3D color spaces. These methods have a pixel-based processing scheme in which for any given pixel, rules are investigated to decide on the class of that pixel. Kovac et al. [27] proposed a method of explicitly defined boundary model using RGB color space in two of daylight and flashlight conditions which has been reutilized in [28]. Here, a pixel is skin in uniform daylight condition if it satisfies R>95, G>40, B>20, max{R,G,B}-min{R,G,B}>15, |R-G|>15, R>G and R>B conditions. In flashlight illumination condition however, a pixel is assumed to b skin if R>220, G>210, B>170, |R-G|<15, R>G, R>B. In [29,30], different elliptical boundaries are estimated based on the fact that skin locus in CbCr is similar to ellipse, and then in evaluation, only pixels which are surrounded by the ellipse are considered as skin. Of course, before segmentation, a preprocessing is performed to enhance the elliptical shape and robustness. For all of these algorithms, the efficiency is totally depending on the rules, training set, and imaging conditions. They are very fast in processing but not reliable for most applications.

There is no explicit definition of probability density function in non-parametric statistical techniques. Single histogram based Look-UP-Table (LUT) model is a common approach in which the distribution of skin pixels in a particular color space is obtained using a set of training skin pixels. Considering RGB as a color space with finest possible resolution (i.e. 256 bins per each channel), the 3D RGB histogram (or LUT) is constructed from 256*256*256 cells each representing skin probability of one possible $R_i G_i B_i$ value. The learning process is though simple, populating the histogram requires massive skin dataset plus huge storage space. The final probability for each possible $R_i G_i B_i$ value is obtained using:

$$P(R_i G_i B_i) = \frac{No.\ of\ occurance\ \ of\ R_i G_i B_i}{total\ counts} \quad (1)$$

This huge storage requirement is histogram model's blemish which is addressed by using both coarser bins and 2D color models. Jones et al. [31] employed this technique for person detection.

In contrary to above LUT approach, the Bayesian classifier considers two histograms of skin and non-skin pixels. In fact, due to the overlap between skin and non-skin pixels in different color spaces (Jones et al. [31] in their study showed, 97.2% of colors which occurred as skin also occurred as non-skin), in above equation, $P(R_i G_i B_i)$ is a conditional probability in which it is already assumed that the observed pixel belongs to skin class. Choi et al. [32] built a Bayesian classifier based on YCbCr color space using MAP (Maximum A Posteriori) for priori estimation. In [33], authors found the adequate number of quantization levels by minimizing an objective function which is the summation of false acceptance rate and false rejection rate in accordance with $2^k$ number of bins of the two histogram.

Parametric statistical models such as single Gaussian models (SGMs), Gaussian mixture models (GMMs), cluster of Gaussian models (CGMs), Elliptical models (EMs), etc are developed to compensate LUT shortcomings. In addition, they generalize very well with a relatively smaller amount of training set [23,34]. Care should be taken in using these models; goodness of fit is a very important parameter in such techniques which specifies how effectively PDF simulates the real distribution [35]. In SGM, there should be a smooth Gaussian distribution around the mean vector. However, in general conditions, the distribution is more sophisticated.

In skin detection, ANNs (Artificial Neural Networks) have been utilized for different purposes and structures. In illumination compensation, dynamic models, in combination with other techniques and direct classification, variety of ANNs such as MLP, SOM, PCNN, etc are exploited. Al-Mohair et al. [36] employed simple 3-layer MLP classifier using different color spaces and different number of neurons in hidden layer. Among common color spaces, they concluded that YIQ gives the highest skin and non-skin separability. In [37], Phung et al. proposed a multi stage technique characterized by several parallel MLPs trained using committee machine. In [38], MLP is trained using back-propagation algorithm to provide an optimized decision boundary and neural network is used for both interpolating skin regions and skin classification.

Hyper-spectral cameras provide useful discriminatory data for biometric applications. Skin detection using imaging devices with hyper-spectral capabilities are not common approaches as they are only applicable using expensive equipments and in particular conditions, though hyper-spectral imagery offers a distinct advantage and dramatically reduce false alarms while maintaining a high detection rate. Recently, Kidono et al. [39] employed a multiband camera to simultaneously obtain seven spectral images to perform pedestrian detection. Suzuki et al. [40] also proposed a method based on the subtraction of two NIR images whose central wavelengths are 870 and 970 nm. Dowdall et al. [41] also proposed a method of skin detection based on the idea that Human skin exhibits an abrupt change in reflectance around 1.4 mm. This phenomenology was employed by taking a weighted difference of the lower band near-IR image and the upper band near-IR image which increases the contrast between human skin and the background in the image. After a simple thresholding, the binary image undergoes a series of opening and closing morphological operations.

Using online information of the image or sequence of frames has been exploited as an effective idea to counteract non-uniform illumination to some extent. Adaptive models are developed in an effort to present models which are calibrated to given inputs. In some cases, previously defined models are tuned for specific conditions i.e. the background, imaging equipment, lighting conditions and even subject of the image. This approach reasonably yields to high detection rate with the cost of loosing generality. In second category of adapted models, an statistic model is updated based on global information of the image. Yang et al. [42] integrated an ANN system to dynamically configure a Gaussian model. The method is based on the fact that distribution of a skin color depends on the luminance level in the Cb-Cr plane. Thus, firstly, according to statistics of skin color pixels, the covariance and the mean value of Cb and Cr with respect to Y channel is calculated, and then it is used to train a neural network which gives a self-adaptive skin color model based on an online tunable Gaussian classifier. In some dynamic approaches, a model is reconfigured based on local information extracted from pre-detected features. Ibrahim et al. [43] employed VJ face detector [44] to segment face regions. Based on face information, the exact boundary of an explicit method in YCbCr is obtained. Using this dynamic threshold, the skin regions all over the image is obtained. However, the performance of the method strongly depends on the accuracy of the face detector. In addition, the method is only useful for images with at least one face.

Some authors have explored different combination of former methods. For example, Zaidan et al. [20] incorporate SAN (segment-adjacent nested technique) under BP ANN and grouping histogram technique under Bayesian method to detect skin regions. SAN is a procedure of generating RGB string value (ranging from 0 to 255255255) of individual R,G and B channels of a pixel. Then, using 3*3 sliding windows, a nested vector of these strings are constructed to be used in neural network. Grouping histogram is algorithm of building skin probability map based on Bayesian rule.

In order to address the shortcomings of pixel-based detectors, recently, variety of spatial based (diffusion based) methods have been developed. The strategy in these techniques is extraction of initial skin seed by means of a precision oriented pixel based detector and annex other skin pixels to the seed. In fact, in ordinary photos, the idea is based on the observation that skin regions are associated with several pact blobs. Thus, it would be reasonable to detect some high probable skin points and then use spatial relationship between pixels to propagate and cover other skin points. Ruiz-de-solar et al. [45] proposed a method of controlled diffusion. In seed generation stage, they utilized a GMM with 16 kernels in $YC_bC_r$ color space and then the final decision about the pixel's class is taken using a spatial diffusion process. In this process, if Euclidean distance between a given pixel and a direct diffusion neighbor (already skin pixel) is smaller than a threshold value, then the propagation occurs. Also, the extension of the diffusion process is controlled using another threshold value, which defines the minimal probability or membership degree allowed for a skin pixel. This process works well in regions where the boundary of skin and non-skin pixels are sharp enough unless the leakage occurs. The effectiveness of this method is questionable as in most conditions, being near a skin pixel is not a concrete clue to be used. Also, using GMM with any number of kernels and any threshold is not reliable for finding seed points.

Abdullah-Al-Wadude et al. [46] also proposed an algorithm uses color distance map (CDM); a gray scale image robust against variations in imaging conditions, and an algorithm based on the property of flow of water which uses spatial analysis to extract skin blobs. Recently, in order to overcome leakage issue, Kawulok et al. [47,48] proposed an energy based approach in which the probability of pixels are utilized to determine the skinness. This method is based on cumulative propagation which is less susceptible to leakage. Another work by the same author [49] is a diffusion procedure based on a distance transform for propagating in a combined domain of skin probability, luminance and hue. After initial seed extraction, a shortest route from the seed to every pixel is determined. Then they used a combined domain to fuse the information of skin probability map and spatial analysis.

In the present work, the idea of spatial analysis has been exploited from different points of view. The extraction of the initial seed, the overall algorithm, the diffusion function, features, diffusion procedure and combination of all available tools are all different from former works.

III. THE PROPOSED ALGORITHM

*A. Overall approach*

The overall approach in the proposed algorithm is discussed in this sub-section and then in the following sub-sections, each part of the algorithm will be elucidated in detail. The data flow of the proposed system is depicted in Fig. 2. In the training phase, a huge number of skin pixels are extracted from the prepared database. Using these pixels, the distribution of the skin cluster is determined in YCb, YCr and CbCr 2D color spaces. This will result in a plot of 3 density maps. These density maps are then utilized to obtain

2 polygons in each of color spaces. These polygons divide each of color spaces into 3 non-overlapping regions. In fact, they specify the boundary between pixels which are most probably associated with human body, pixels which can be a human skin in some conditions and pixels which are not probably related to skin. In evaluation step, for each input image, using estimated polygons, the color image is converted to a ternary image. In addition, Otsu segmentation is also performed on the image in order to generate homogenous regions of the image. Also, canny edge detector with relatively high threshold is employed to obtain the strong edges in input image and avoid leakage in diffusion step. Using the initial seed (of ternary image), segmented images in 6 different color components and the so called edge map of the image, the skin pixels of the color image will be specified.

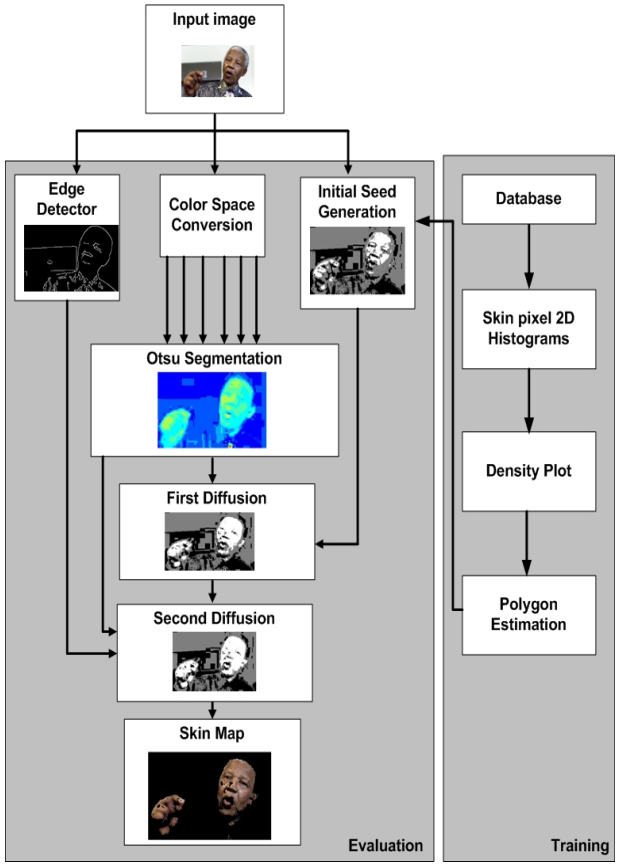

Fig. 2. Proposed Algorithm

### B. Training phase

In order to construct the skin cluster model, a database of approximately 63 million skin pixels was exploited. These pixels were randomly chosen with particular attention to preserve the generality of the algorithm. Thus, images were selected from human pictures in different lighting conditions, variety of ethnicities, disparate ages, sexes, etc. This database of skin pixels is employed in order to obtain human skin cluster in $YC_bC_r$ color space. It is claimed that the optimum performance of a skin classifier is not dependant on the employed color space [4]. However, in this paper, the skin cluster is obtained in $YC_bC_r$ color space for some reasons First, compared with RGB, redundancy of its channels is much lower and its components are more independent [23]. Also, it makes it easy to use it in many applications without a need to transform the image into another color space. Thirdly, as it has been statistically analysed in [50], the performance of $YC_bC_r$ is superior compared to others when using the same method for different color spaces. Also, one advantage of $YC_bC_r$ compared to many other color spaces like HSV and TSL is that it can be obtained via a linear transformation from commonly used RGB format.

Based on collected data in $YC_bC_r$, three sets of pixels are defined. These sets are calculated in accordance with intensity heat maps (density plots). Each map is divided into 3 non-overlapping partitions. The boundary of each sector is estimated by a polygon in order to preserve the simplicity of the algorithm. For each map, 2 polygons are estimated. The inner polygon represents the boundary between the pixels which have been observed with high frequency and the pixels which are not highly repeated, but they can still be part of skin in some conditions. The second boundary separates pixels which have been observed with low frequency and pixels which are not observed ever. This has to be carried out for each of the three maps. Based on above definitions, set $T_1$ contains pixels that are simultaneously located inside all three inner polygons. $T_2$ contains pixels which are not included in T1 and they are inside the outer polygons. Finally, set $T_3$ is filled with the rest of pixels. By this partitioning, dealing with nonlinear illuminations will be much easier.

### C. Initial seed generation

The first step in evaluation of the proposed method is to generate ternary image. Here, contrary to other methods, the image is not directly classified into a skin and non skin pixels. Based on the so called sets, the image will be segmented to 3 groups of pixels. In fact, there are lots of uncertain pixels which deciding that whether or not they belong to skin is difficult mainly due to illumination, imaging device effects and other reasons. Therefore, instead of roughly discard or accept them, they will be put aside for further processing steps based on spatial analysis.

In pixel-based processing, each pixel of the input image is classified according to above three defined sets. $T_1$ pixels are set white (255), $T_3$ pixels are set black (0) and $T_2$ pixels are set gray (128). By this, a ternary image is obtained. In this case, some skin regions are detected easily as they are not affected by non linear illumination, and they more or less will be specified by most methods (they are located in $T_1$ set). Gray pixels are the ones that no strict decision can be made for them; some of them are truly skin pixels, and some are not. It should be noted that black pixels are not yet discarded. They might be annexed into skin group of pixels later in diffusion stage, if their score in diffusion function would be enough. In fact, during diffusion, black pixels are less likely to be diffused than gray pixels.

Subsequently, except for bordering pixels, and skin pixels which have been verified before ($T_1$ pixels), each pixel of the ternary image is processed in a neighbour based algorithm. To that end, A 3*3 and a 5*5 neighbourhood is considered for each pixel and a score is calculated based on the number of $T_1$, $T_2$ and $T_3$ type pixels in its neighbours, and by using a set of rules on the number of pixels in neighbours. The score is calculated by (1), in which $\zeta$ is the total score, K is a constant, T is the score in the 3*3 neighbour and $\phi$ is the same in the 5*5 neighbour.

$$\xi = K \times T + \Phi \tag{2}$$

Decision is made based on the value of $\zeta$. Two thresholds have been statistically determined in order to decide whether a pixel is skin or not. A pixel with score lower than $th_1$ is considered black and higher than $th_2$ is considered white (hysteresis thresholding). The output of this stage is another ternary image that for some of its pixels, decision has not been made yet (skin or not skin). It has been observed that for many other skin pixels with considerable number of skin pixels around them, they have not been set white yet, and this may be due to illumination condition, a shadow, moustache, etc. The proposed neighbour based procedure compensates these effects to some extent. Also, isolated pixels mainly caused by noise will be eliminated.

Fig. 3 shows the result of employing the proposed method in finding skin pixels in particular skin region. In this stage, there is no intention in finding all of skin regions, but the goal is firstly, finding probable complexion pixels in each possible region (even one point!) and secondly, determining other probable skin regions. Deliberately, images with different illumination conditions, different ethnicities, facial expressions and even age are selected to corroborate the validity of the system. In the following, each of the skin pixels will be selected and exploited to determine the essence (skin or non-skin) of other pixels in vicinity regions. Initial seed points are extracted based on white points in final ternary image.

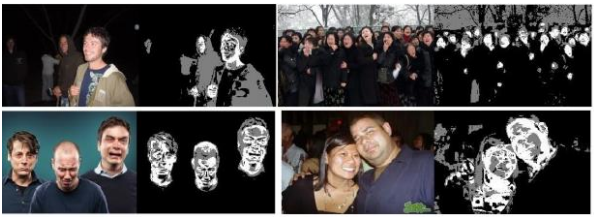

Fig. 3. A number of selected images from evaluation database and their corresponding ternary conversion

### D. Otsu segmentation

Otsu method of image segmentation [51] is a nonparametric and unsupervised method of automatic selection of an optimal threshold. This method is optimum as it maximizes the between-class variance. This method is also computationally important as it is only based on 1-D histogram of an image. The procedure is very simple just utilizing the zeros and the first order cumulative moments of the histogram. The Otsu method is often employed for segmentation of gray level images. However, in this paper, it is utilized for images in different color spaces and color components. In Fig. 4, the result of performing Ostu segmentation on several color channels is depicted. Starting with the main image in the left column, the ternary image is the second image and then from top to bottom it goes by the results of performing Otsu segmentation technique using Y, Cb, Cr of YCbCr color space. H and S components of HSV, I and Q of YIQ and X of XYZ color space are presented in second column. In the third one, the result of applying segmentation on the Y and Z components of XYZ color space, Cm and K of CmYK and u of Luv is depicted and in last column v of Luv space, a and b of Lab and c, h of Lch color spaces is represented.

Experimentally, it has been observed that in most of images, there are certain numbers of skin pixels that for different reasons, they have been excluded in the initial segmentation. Unfortunately, these pixels are not a few in number. Thus, a method should be designed to distinguish these pixels. An approach is proposed here. Looking at Fig. 4, it is clear that for some of the color channels, human faces (all 4 faces in the figure) are encompassed in one or two classes (of segmentation) while it is not the case for other channels. For instance, in 3$^{rd}$ image of the left column, classes in face regions are too varied to be used for this purpose. But, for two lower images, the situation is different and face regions are almost homogenous. This means applying Otsu method on some color spaces can be leveraged in order to identify homogonous regions to some extent. Furthermore, fusion of the result of different color components will strength the power of the classifier that designed based on this approach. So, approximate homogenous regions are estimated using fusion of Otsu segmentation in Cb, Cr, I, H, u, a, and c. It should be noted that these color components are not utilized with the same weight in the diffusion part, i.e., some of them are more robust considered with higher impacts on the diffusion score.

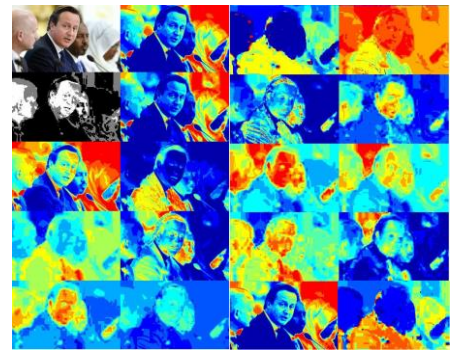

Fig. 4. Otsu clustering technique in different color components

### E. Edge Detection for Leakage Avoidance

There is a tangible problem in diffusion methods and the concept of diffusion. Diffusing, in a general definition, means propagation from places, things, pixels, etc. that are relatively dense, to lower dense regions. However, care should be taken to avoid leakages. In skin segmentation, leakage happens when a skin pixel find an opportunity to diffuse into non-skin regions. Obviously, this will lessen the performance of the system. In order to address this issue as much as possible, canny edge detector is put forward. The edge detector is designed to find relatively strong edges of the image. This will be exploited in diffusion part in order to bridle undesired propagation. In fact, when diffusion is performed, the edge detected image functions as a barrier and it would not let any pixel to propagate through it. Obviously, the performance of the edge detector will be decisive, thus canny edge detector is employed. Small particles and edges are not required as they can disfigure the final image. Thus, there should be a compromise on the edge threshold to both preserve the effectiveness and vital role of edge detector and also avoid unnecessary detail edges. A proper threshold value is an empirical choice may vary from image to image. The final results of applying edge detection on some input images, by choosing a unique threshold value is shown in Fig. 5. The proposed leakage avoidance method is effective in many situations; however, there are certain conditions in which the edge detector is not perfectly adopted. In most of these cases, the diffusion score will stop the propagation process by itself. Nevertheless, the final result (skin segmented image) will not differ significantly with these limitations.

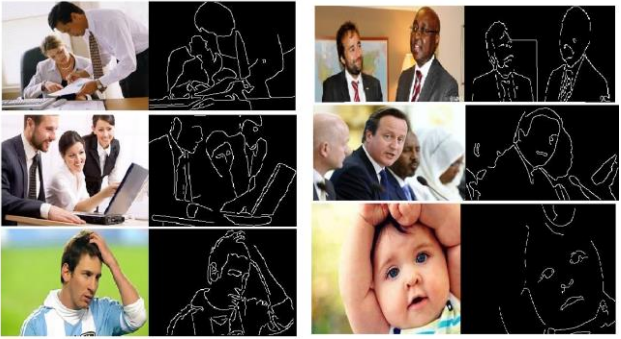

Fig. 5. Edge detection in evaluation phase

### F. Diffusion

The idea of diffusion is propagation from certain skin pixels to other non-initially-specified skin regions. This is done in two separate but similar stages. The approach in both stages is that the image is scanned from the top left pixel to the final bottom right pixel to search for any skin pixel in the initial seed. Finding any skin pixel, propagation starts. When propagation is finished for that particular skin pixel, the procedure continues for other skin pixels till reaching the end of the image.

The proposed propagation scheme is presented in Fig. 6 in which white points show skin seed and red dots represent diffusion direction. For each candidate white point, the process starts with nearest neighbor pixels and it continues to surrounding image pixels. In fact, for each seed point, 36 virtual lines are considered i.e. the 2D space around the master pixel is segmented into 36 regions using virtual lines with 10° resolution. The diffusion is performed on each of these lines. Beginning with the line in 0°, the first pixel to investigate is the rightmost pixel of that master pixels. Then, other pixels in that direction will be processed until propagation stops for some reasons. When processing of one line is finished, the tilt of the virtual line will be increased by 10° and then propagation commences in that direction. The only difference between two diffusion schemes is that the first diffusion is considered to add many other probable skin pixels into initial seed, and in the second diffusion; the added pixels, themselves can be leveraged for finding other skin pixels. In second diffusion process, the new diffused pixels will not be utilized for starting propagation again.

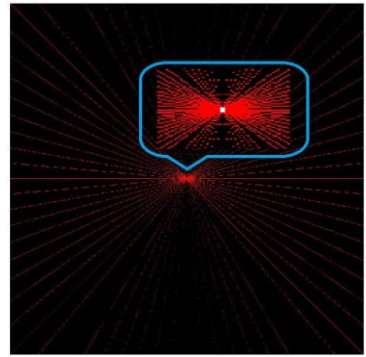

Fig. 6. Virtual lines of diffusion scheme

The criterion used to decide whether to diffuse into specific pixel or reject it is based on the concept of nonlinear diffusion function. For each under-test pixel (UTP), if it is white already, diffusion continues to the next pixel in the same direction, if it is part of the edge map, propagation is terminated in that direction, and if it does not satisfy any of these conditions, then the diffusion function (S) will be calculated and based on the result, the decision will be taken.

$$S = \sum w_i f_i \qquad (3)$$

The score of diffusion (S) for a certain UTP pixel depends on several factors ($f_i$) with unequal weights ($w_i$) and this is due to the fact that some of those factors are more effective and decisive. Those factors include whether the UTP is a gray or black pixel (in corresponding ternary image), and also the difference between the master pixel's class number (after Otsu segmentation) and UTP's one in all of 6 specified color channels. This is logically explainable as when a black pixel exists near a white pixel and both of them are included in a homogenous region, then the black one is also probably skin. The same fact is true for gray pixels (even with lower different class cost).

## IV. EXPERIMENTAL RESULTS

There are several skin detection datasets which can be used both for training and evaluation of algorithms. However, images lack high quality, and ground truths are not accurate in most cases. Ground truths (GT) of Compaq [31] which is one of the earliest skin datasets and other skin datasets also suffered from a problem, which for many pixels taking decision on the identity of them (even manually) is not straightforward. These pixels are located in acute regions such as the boundary of skin and non-skin pixels, head holes, lips, eyebrows, etc. In addition, exact discrimination of skin and non-skin pixels in some regions is very time-consuming and un-useful task. In order to address shortcomings of other datasets, a through skin and face analysis database has been provided in order to investigate the performance of the propose skin detection systems. Skin Detection Database (SDD) includes more than 20,000 (skin and non-skin) color images captured in different conditions, from variety of people all around the world to avoid interference of any external factor in evaluation results. Images were collected by surfing online web images and several face detection databases. Several SDD images are represented in this paper. This database is set for public use and it is the biggest standard and relatively precise dataset for evaluation and training tasks.

In SDD, pixel intensities in GT images are grouped into 3 sets. Those in the category of "Red" are associated with skin pixels, black points represent non-skin pixels and blue ones are unknown (or inconsiderable) pixels. Red points are related to the pixels which are exactly extracted from skin region and black pixels are all associated with non-skin regions. Uncertain pixels are marked with blue color and they will be ignored in evaluation process. The ground truth images were all manually annotated (rather than using a semi-automatic procedure like many other databases) to ensure high precision.

Both quantitative and qualitative results are illustrated here. Figs. 7 and 8 represent the result of performing skin segmentation on a set of random images using the proposed method. As it is observed, all images are successfully segmented, though there are possible false rejections and detections. The woman's hair in Im.3 and man's glass in Im.6 are not miss-classified. In fact, in the proposed method, if there would be a non-skin object particularly with low thickness nearby a skin region, then the region will be probably diffused. For example, eyes, lips, eyebrows, glasses and sometimes necklaces and bracelets are affected. Of course, it depends on both detected or non-detected edges and scores in Otsu segmentation classification. But from practical point of view, it does not cause problem in many applications such as face detection/tracking/recognition and image filtering where skin segmentation is often utilized for reduction purposes. Im.2 is an example to show how the proposed method is capable of dealing with ethnicity problem and images with difficult face postures. Im.9 is also another example of challenging imaging conditions, illumination and ethnicity. Im.9 shows how the proposed skin detection is not affected by face and body orientations.

Images 11 to 16 in Fig. 7 lack complex background but are good samples for measuring how effective is the algorithm in dealing with non-linear illumination. For Images 17-29, some photos are segmented with remarkable precision and recall rates, however for images such as Im.18 and 19, the result is not satisfactory mainly due to the excessive red component in conjunction with little but tangible shadow in both faces in the Images.

Fig. 8 includes more multifarious images. As it is observed, skin regions in all of the images are detected strikingly. The second point is that not only the system is able to deal with different skin tones, but it can be also exploited in crowded images with unlimited number of people and unconstrained size of faces. There exist 4 faces in Im.34 and Im.37, 11 faces in Im.35, 8 faces in Im.36, 22 ones in Im.38, 10 faces in Im.39 and 6 faces in Im.40 which all of them are segmented with acceptable rates clarifying the fact that the proposed system is a well-choice for filtering purposes in front-end of face detectors. Combination of this method and state-of-the-art face analyzers such as a texture-based Viola-Johns face detector [44] will yield to a very efficient system. For images number 46-55 in Fig. 8, variety of devices and illumination conditions have been involved. Results demonstrate the merit of the proposed algorithm in dealing with such challenges. Moreover, for images 56 to 63 captured in disparate outdoor conditions; some have backgrounds with skin-like pixels, some of them are old images lacking enough quality and some contain several faces; and in overall, the system has processed them effectively.

Quantitative results are also examined. The performance of the system is evaluated by using standard metrics that is common in classification problems. Precision and recall are two other important metrics can be used to evaluate skin detection algorithms. In skin segmentation, precision and recall are often in trade-off with each other and F-score is a useful tool to describe their relation. Table. I compares the performance of different works. As it is observed, the performance of the proposed method is impressive in compare with other methods. Evaluation has been performed based on a number of random images in SDD for all methods. Also, Fig. 9 shows a set of images extracted from previous published works accompanying with the results of applying the proposed method. In the figure, Images are from [52-59] respectively. In the triple images of 1-8, the left one is the original photo, the middle is extracted from other papers and the right image is based on proposed algorithm.

V. CONCLUSION

In this paper, an efficient skin segmentation algorithm that can effectively classify each pixel of an arbitrary image into two classes of skin and non-skin one is proposed. The system consists of several steps beginning with initial seed generation. In this stage, the algorithm separates the pixels into three set of the high probable skin pixels, high probable non-skin pixels and the rest of pixels. The most probable skin ones will be utilized in subsequent steps to enhance the detection rate. The edge map of the input image by using canny edge detector is prepared in conjunction with Otsu segmented images in different color space components. The former is to avoid leakage in diffusion and the latter is to virtually segment image into homogonous regions. The final stage is diffusion of the initial seed. The system has been evaluated by using SDD which is an exhaustive standard skin dataset, with manually annotated ground truths. The merit of the proposed system was both quantitatively and qualitatively represented in compare with several former works. There are three important points which are under consideration to expand this study. First is the performance of initial seed generator that is very important and its improvement will directly impact the entire system. Secondly is designing of the diffusion function such that several other factors can be included to boost the system's performance including texture and skin probability map. And finally, in order to provide a real-time system, possible FPGA implementation of the proposed system is under investigation.

TABLE I.

PERFORMANCE OF DIFFERENT METHODS

| Method | Precision | Recall | F-score |
|---|---|---|---|
| Vadakkepat el al. [7] | 0.5411 | 0.4046 | 0.4630 |
| Hsu et al. [60] | 0.4996 | 0.6796 | 0.5758 |
| Albiol et al. [4] | 0.4004 | 0.7331 | 0.5178 |
| Sagheer et al. [61] | 0.4470 | 0.2894 | 0.3514 |
| Gang et al. [62] | 0.2809 | 0.7313 | 0.4058 |
| Huang et al. [63] | 0.3688 | 0.7166 | 0.4868 |
| Pai et al. [64] | 0.1834 | 0.2866 | 0.2236 |
| Wu et al. [65] | 0.4920 | 0.6286 | 0.5518 |
| Zhu et al. [66] | 0.3067 | 0.7097 | 0.4283 |
| Thakur et al. [67] | 0.4526 | 0.6879 | 0.5460 |
| Yutong et al. [68] | 0.4726 | 0.7121 | 0.5681 |
| Qiang-rong et al. [30] | 0.5508 | 0.5634 | 0.5570 |
| Subban et al. [69] | 0.2986 | 0.5509 | 0.3872 |
| Chen et al. [70] | 0.2936 | 0.7080 | 0.4150 |
| Anghelescu et al. [71] | 0.1366 | 0.7331 | 0.2302 |
| Wang et al. [72] | 0.5748 | 0.5465 | 0.5603 |
| Proposed | 0.85 | 0.71 | 0.7737 |

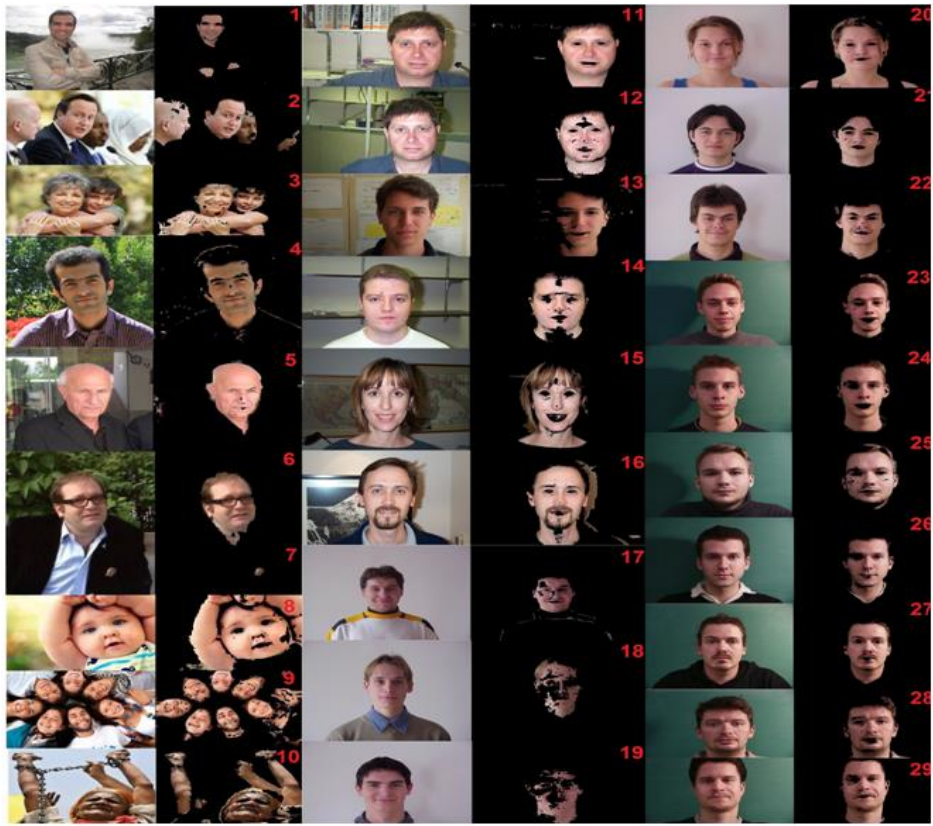

Fig. 7. Images from Web, Caltech, CVL and IMM

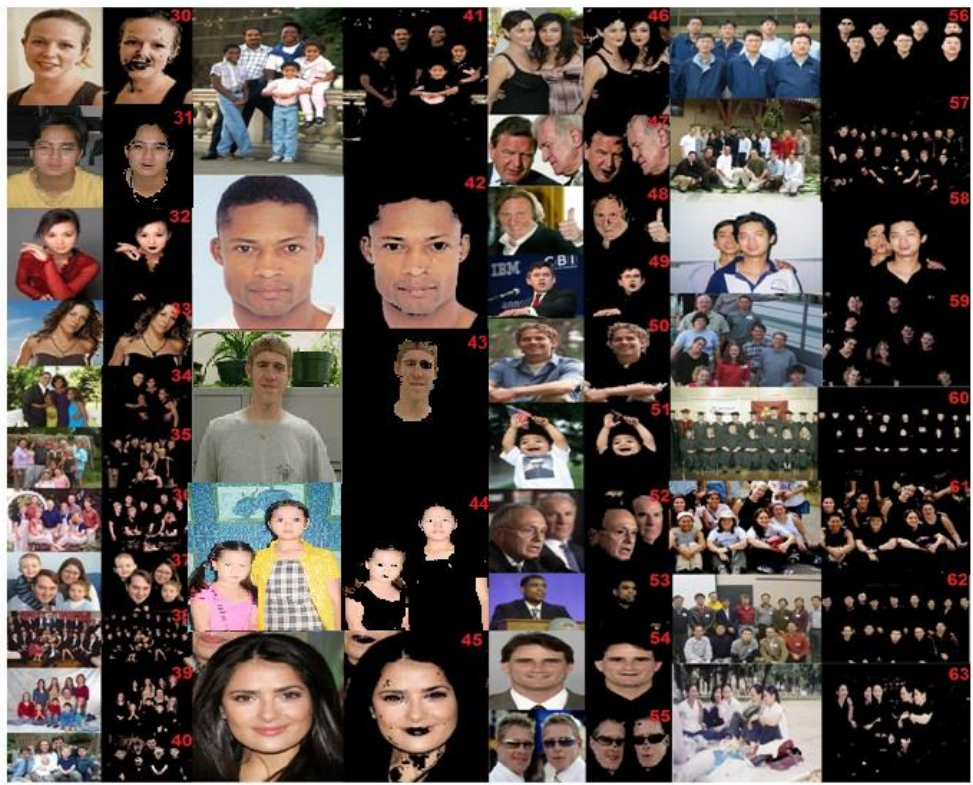

Fig. 8. Images from Pratheepan, LFW and Bao

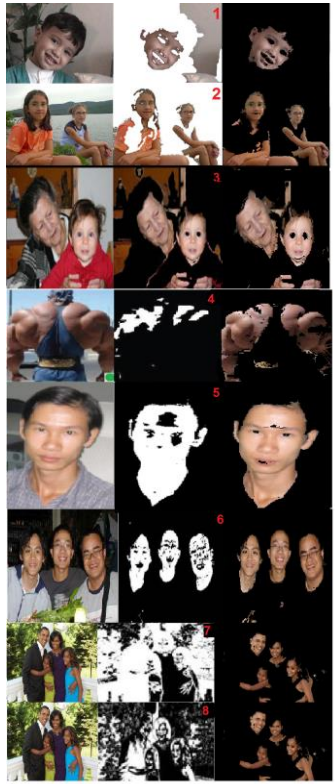

Fig. 9. Comparing the performance of proposed method with previous works